\pdfoutput=1
\documentclass[11pt]{article}
\usepackage{ACL2023}

\usepackage{times}
\usepackage{latexsym}
\usepackage[T1]{fontenc}
\usepackage[utf8]{inputenc}
\usepackage{microtype}
\usepackage{inconsolata}
\usepackage{multirow}
\usepackage{amsmath}
\usepackage{graphicx}
\usepackage[most]{tcolorbox}
\usepackage{color,soul}
\usepackage{pifont}         

\usepackage[table]{xcolor}  
\usepackage{booktabs}       
\usepackage{array}          
\usepackage{colortbl}       

\definecolor{greenhl}{RGB}{181, 215, 155}
\definecolor{bluehl}{RGB}{160, 160, 200}
\definecolor{purplehl}{RGB}{150, 120, 180}
\definecolor{redhl}{RGB}{220, 120, 120}     
\definecolor{orangehl}{RGB}{255, 165, 0}    
\definecolor{yellowhl}{RGB}{255, 255, 153} 
\definecolor{lightblue}{RGB}{173, 216, 230}
\newcommand{\cmark}{\textcolor{green}{\ding{51}}} 
\newcommand{\xmark}{\textcolor{red}{\ding{55}}} 

\title{ACE-ICD: Acronym Expansion As Data Augmentation For Automated ICD Coding}

\author{
  \textbf{Tuan-Dung Le\textsuperscript{1,2}},
  \textbf{Shohreh Haddadan\textsuperscript{1}},
  \textbf{Thanh Q. Thieu\textsuperscript{1,2}}
\\
\textsuperscript{1}Moffitt Cancer Center and Research Institute, \textsuperscript{2}University of South Florida
\\
  \texttt{\{tuandung.le, shohreh.haddadan, thanh.thieu\}@moffitt.org} 
\\
}

\begin{document}
\maketitle
\begin{abstract}
Automatic ICD coding, the task of assigning disease and procedure codes to electronic medical records, is crucial for clinical documentation and billing. 
While existing methods primarily enhance model understanding of code hierarchies and synonyms, they often overlook the pervasive use of medical acronyms in clinical notes, a key factor in ICD code inference. 
To address this gap, we propose a novel effective data augmentation technique that leverages large language models to expand medical acronyms, allowing models to be trained on their full form representations.
Moreover, we incorporate consistency training to regularize predictions by enforcing agreement between the original and augmented documents.
Extensive experiments on the MIMIC-III dataset demonstrate that our approach, \textbf{ACE-ICD} establishes new state-of-the-art performance across multiple settings, including common codes, rare codes, and full-code assignments.
Our code is publicly available \footnote{\url{https://github.com/LangIntLab/ACE-ICD}}.
\end{abstract}

\section{Introduction}
Assigning standardized codes based on the International Classification of Diseases (ICD\footnote{\url{who.int/standards/classifications/classification-of-diseases}})
known as ICD Coding is essential for efficient medical record management, accurate billing processes, and streamlined insurance reimbursements \cite{park2000accuracy, sonabend2020automated}.
However, traditional ICD coding relies on manual effort, making it time-intensive and error-prone driving the development of automated coding methods.

\begin{table}[h]
    \centering
    {\fontsize{9}{10}\selectfont
    \setlength{\tabcolsep}{2pt}  
    \renewcommand{\arraystretch}{1.1}  
    \begin{tabular}{p{4.8cm}cc}
        \hline
        \rowcolor{black!10} \multicolumn{3}{c}{\textbf{Discharge Summary}}\\ 
        \hline
        \multicolumn{3}{l}{... history of present illness: ortho hpi: 86m w/ severe}\\
        \multicolumn{3}{l}{b/l \colorbox{yellowhl}{oa}, admitted to ortho for sequential bilateral \colorbox{bluehl}{tka} ...} \\
        \multicolumn{3}{l}{icu hpi: 86 y/o m with pmhx of arthritis, bph \& } \\
        \multicolumn{3}{l}{osteoporosis s/p elective right \colorbox{bluehl}{total knee replacement} ...} \\
        \multicolumn{3}{l}{past medical history: osteoporosis \colorbox{lightblue}{anemia} (family h/o}\\ 
        \multicolumn{3}{l}{g6pd deficiency) bph \colorbox{yellowhl}{osteoarthritis} cataracts ...}\\
        \multicolumn{3}{l}{empiric vancomycin and ceftriaxone for possible \colorbox{greenhl}{uti}}\\
        \multicolumn{3}{l}{were initiated ...}\\

        \hline
        \rowcolor{black!10} \multicolumn{1}{c}{\textbf{Label}} & \textbf{KEPT} & \textbf{ACE-ICD} \\
        \hline
        \colorbox{greenhl}{\textbf{599.0}} urinary tract infection, site not specified & \xmark & \cmark \\ 
        \hline
        \colorbox{yellowhl}{\textbf{715.36}} osteoarthrosis, localized, not specified whether primary or secondary, lower leg & \xmark & \cmark \\ 
        \hline
        \colorbox{bluehl}{\textbf{81.54}} total knee replacement & \cmark & \cmark \\ 
        \hline
        \colorbox{lightblue}{\textbf{285.9}} anemia, unspecified & \cmark & \cmark \\ 
        \hline
        ... & ... & ... \\
        \hline
    \end{tabular}
    }
    \caption{Example predictions from the MIMIC-III-full dataset (HADM\_ID = 108519) using KEPT\cite{Yang2022KnowledgeIP} and our ACE-ICD models.}
    \label{tab:comparison}
\end{table}

\begin{figure*}[t]
  \includegraphics[width=\textwidth]{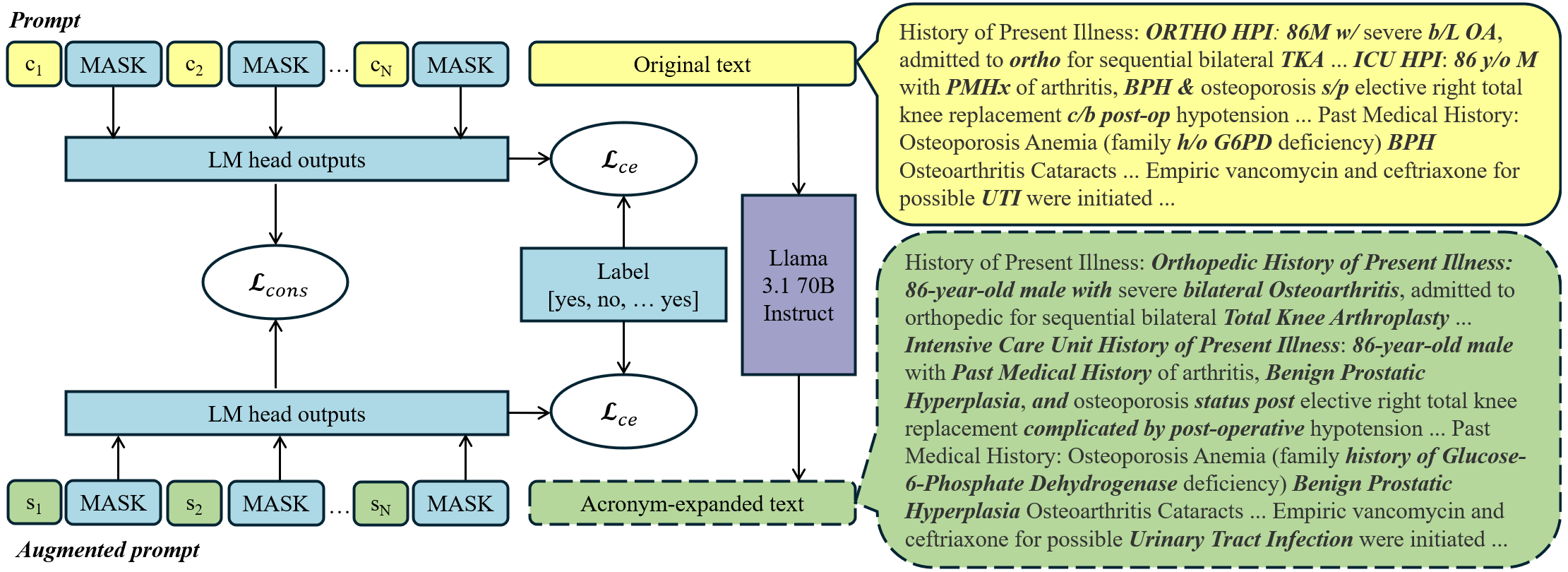}
  \caption{Our training pipeline incorporating acronym-expanded data augmentation and consistency training.}
  \label{fig:pipeline}
\end{figure*}

Accurate ICD code assignment, whether manual or automated, requires a comprehensive understanding of clinical notes, which include detailed information such as symptoms, diagnoses, and test results. However, healthcare professionals often rely on acronyms and abbreviations to reduce documentation effort \cite{amosa2023clinical}. This shorthand introduces significant ambiguity, making it difficult for both human coders and language models to interpret the clinical text correctly. To better understand the impact of acronyms on ICD code assignment, we analyze expert-annotated evidence spans from the MDACE dataset \cite{cheng-etal-2023-mdace}, which is derived from MIMIC-III \cite{johnson2016mimic}. Our analysis shows that 22\% of the evidence spans either consist entirely of acronyms and abbreviations or include at least one such term, highlighting that acronyms often carry useful information for assigning ICD codes.
Previous studies improved model understanding of medical acronyms by leveraging their appearance in code synonyms \cite{yuan-etal-2022-code, Yang2022KnowledgeIP, gomes-etal-2024-accurate}. 
However, they still fail to predict codes which contain acronyms in their synonym description. 
Table \ref{tab:comparison} shows several instances of this shortcoming in previous models.
These failures underscore the need for new approaches to effectively address medical acronyms to consequently improve the ICD coding system.

In this paper, we propose \textbf{AC}ronym \textbf{E}xpansion for \textbf{ICD} coding (\textbf{ACE-ICD}) to investigate the impact of expanding medical acronyms to full terms on coding models. 
Our system harnesses the strong capability of large language models in disambiguating clinical acronyms \cite{kugic2024disambiguation, liu2024exploring}, as a data augmentation method that expands acronyms using open-source LLM prompting. 
We further apply KL divergence consistency regularization to enforce alignment between predictions from the original and augmented examples.
Finally, we conduct experiments on three standard ICD coding tasks using the MIMIC-III dataset, demonstrating that our approach outperforms previous state-of-the-art methods across all tasks.

\section{Methods}

Previous methods frame the ICD coding task as a multi-label classification problem, as a single clinical note can contain multiple diagnosis or procedure codes. 
Given a clinical note $t$, the task is to assign a binary label $y_i \in \{0, 1\}$ for each ICD code $i$ (where $i=1,2,...,N_c$ and $N_c$ is the total number of ICD codes). 
A label of 1 denotes relevance, while 0 indicates irrelevance.

We followed the prompt-based fine-tuning approach by \cite{Yang2022KnowledgeIP} for the ICD coding task, reformulating the multi-label classification task as a cloze task \cite{schick-schutze-2021-exploiting, gao-etal-2021-making}. 
Specifically, we construct a prompt template by concatenating each ICD code description $c_i$, appending a [MASK] token after each description, and adding the clinical note $t$. 
The prompt $P$ is given by: $P$ = $c_1$ [MASK] $c_2$ [MASK] ... $c_{N_c}$ [MASK] $t$.
The model is trained via a masked language modeling objective to predict “yes” or “no” in each [MASK] position, corresponding to a label of 1 or 0, respectively.
For tasks where $N_c$ is large (e.g., thousands of codes), this approach is typically used as a reranker \cite{tsai-etal-2021-modeling, Yang2022KnowledgeIP, yang2023multi, pmlr-v225-kailas23a}, as including all code descriptions in the prompt is infeasible.
\subsection{Acronym-expansion as data augmentation}
Motivated by capabilities of LLMs in clinical acronym disambiguation \cite{liu2024exploring, kugic2024disambiguation}, we use the open-source Llama 3 model \cite{dubey2024llama} to generate acronym-expanded version of clinical notes, denoted as $t_{a}$, from the original notes $t$.
Due to the considerable length of the MIMIC-III notes, we first split each discharge summary based on the headers identified through automatic section-based segmentation \cite{lu2023towards}. 
We then prompt the instruction-tuned Llama 3 models to generate augmented sections, which are concatenated into the final acronym-expanded note, using the following prompt:

\begin{tcolorbox}[colframe=black!75!white, colback=gray!10, boxrule=0.5mm, halign=flush left,  fontupper=\scriptsize]
\texttt{<|begin\_of\_text|>}

\texttt{<|start\_header\_id|> system <|end\_header\_id>}

\texttt{You are a helpful assistant. <|eot\_id|>}

\texttt{<|start\_header\_id|> user <|end\_header\_id|>}

\texttt{Expand all acronyms to their full forms while preserving all the details in the following paragraph, do not mention the acronyms again. Paragraph: \_\_\_ <|eot\_id|>}

\texttt{<|start\_header\_id|> assistant <|end\_header\_id|>}

\texttt{Here is the paragraph with all acronyms expanded to their full forms: }
\end{tcolorbox}

\subsection{Consistency training}
Inspired by \cite{shen2020simple, wu2021r}, we incorporate a consistency loss into the training objective to encourage the model to generate similar predictions for the original clinical note $t$ and its acronym-expanded counterpart $t_a$.
We first construct a second prompt template using the augmented note $t_a$ and a synonym $s_i$ for each ICD code description collected from UMLS from previous studies  \cite{yuan-etal-2022-code, gomes-etal-2024-accurate}: $P_a$ = $s_1$ [MASK] $s_2$ [MASK] ... $s_N$ [MASK] $t_a$.
The training objective can be written as the following:
\begin{multline}
\mathcal{L} = \frac{1}{2}(\mathcal{L}_{ce}(P, y) + \mathcal{L}_{ce}(P_{a}, y)) \\
+\alpha\mathcal{L}_{cons}(P, P_{a}, y) \nonumber
\end{multline}

where $\mathcal{L}_{ce}$ is cross-entropy loss for masked language modeling, and $\mathcal{L}_{cons}$ enforces consistency by minimizing the bidirectional KL-divergence: 
\begin{multline}
\mathcal{L}_{cons}=\frac{1}{2}(KL[p(y|P)||p(y|P_a)] \nonumber\\
+ KL[p(y|P_a)||p(y|P)])
\end{multline}

\section{Experiments}
\subsection{Acronym expansion evaluation}

\quad \textbf{Dataset.} \quad Several annotated datasets for medical abbreviations and acronyms have been introduced in prior work \cite{moon2012clinical, rajkomar2022deciphering}. To evaluate our zero-shot prompting approach on the MIMIC-III corpus, we utilize the annotated dataset developed by \citet{rajkomar2022deciphering}, which we consider most suitable for our task. This dataset was constructed using a reverse-substitution technique applied to MIMIC-III discharge summaries, resulting in a large-scale, high-quality resource for acronym expansion. As our experiments are also based on MIMIC-III, this dataset offers a consistent and reliable benchmark for evaluation.

\quad \textbf{Metrics.} \quad We perform zero-shot prompting experiments using four instruction-tuned variants of the Llama model. We first report detection precision and recall \cite{rajkomar2022deciphering}, which assess the model's ability to identify abbreviations in text, regardless of whether the corresponding expansions are correct. We then compute total accuracy, defined as the proportion of abbreviations in the gold standard that are correctly expanded to their full forms by the model. To assess the quality of expansions, we consider two evaluation settings. Strict accuracy requires an exact string match between the model-generated expansion and the gold-standard full form. However, as illustrated in Table~\ref{tab:abreviation_examples}, certain expansions may differ lexically yet convey the same meaning. To accommodate such cases, we introduce lenient accuracy, which computes a similarity score based on the normalized inverse edit distance between the generated and reference expansions. Specifically, we normalize the edit distance by the length of the reference and consider expansions with a similarity score of at least 70\% as correct. Although some semantically valid expansions may fall below this threshold, we adopt the 70\% cutoff to balance recall and precision while avoiding acceptance of clearly incorrect outputs.

\quad \textbf{Implementation Details.} \quad To extract abbreviations and their corresponding expansions from model outputs, we employ the Python difflib library\footnote{\url{https://docs.python.org/3/library/difflib.html}}, which aligns two sequences.
\label{acronym_eval}



\subsection{ICD coding evaluation}

We evaluate our methods on three MIMIC-III tasks, following \cite{mullenbach-etal-2018-explainable} for MIMIC-III-50 and MIMIC-III-full dataset splits and \cite{Yang2022KnowledgeIP} for constructing MIMIC-III-rare50, which focuses on the top 50 codes with fewer than 10 occurrences (See Table \ref{tab:dataset}).

\begin{table}[h]
\centering
{\fontsize{9}{10}\selectfont  
    \setlength{\tabcolsep}{4pt}  
    \renewcommand{\arraystretch}{1.2}  
\begin{tabular}{lcccc}
\hline
    \textbf{Dataset} & \textbf{Train} & \textbf{Dev} & \textbf{Test} & \textbf{$N_C$}\\
\hline
MIMIC-III-50 & 8066 & 1573 & 1729 & 50\\
MIMIC-III-full & 47723 & 1631 & 3372 & 8922\\
MIMIC-III-rare50 & 249 & 20 & 142 & 50\\
\hline
\end{tabular}
}
\caption{\label{tab:dataset}
Statistics of the MIMIC-III ICD-9 datasets.
}
\end{table}

\textbf{Metrics.} \quad We evaluate performance using macro-AUC, micro-AUC, macro-F1, micro-F1, and precision@k (k = 5 for MIMIC-III-50, k = 8 and 15 for MIMIC-III-full). 
We determine the optimal threshold for micro-F1 on the development set and report test metrics using the best-performing checkpoint. 
To ensure robustness, we run each experiment with five random seeds and report mean results.

\begin{table*}[t]
\centering
{\fontsize{9}{10}\selectfont  
    \setlength{\tabcolsep}{3pt}  
    \renewcommand{\arraystretch}{1.2}  
\begin{tabular}{lccc}
\hline
    \textbf{Acronym} & \textbf{Expanded form by LLM} & \textbf{Full form} & \textbf{Inverse Edit distance}\\
\hline
pod\#15 & post-operative day \#15 & post-operative day \#15 & 100.0\\
intrabd& intrabdominal &intra-abdominal & 86.67\\
p/w & presented with 	 &presents with & 84.61\\
dvt & deep vein thrombosis & deep venous thrombosis& 81.82\\
 \hline
co 	& complained about& 	 complained of &	 69.23\\
angio & angiography & angiogram& 66.67\\
dec & dead& deceased& 50.0\\
x3 & three& three times & 45.45\\
\hline
p.a. & personal assistant& physician assistant & 68.42\\
pms & previous medical symptoms. & premenstrual syndrome& 28.57\\
vma & visual motor assessment & vanillylmandelic acid & 0.0\\
opa & office of personnel administration& oropharyngeal airway &-25.0\\
\end{tabular}
}
\caption{\label{tab:abreviation_examples}
Examples of expanded abbreviations, their correct full-forms and the value of the length-normalized inverse edit distance. The threshold is considered 70\%, thus the expanded terms with a lower threshold are considered incorrect in our evaluation.}
\end{table*}

\textbf{Implementation Details.} \quad We initialize our model with two pre-trained language models: KEPTLongformer\footnote{\url{https://huggingface.co/whaleloops/keptlongformer}} and KEPT-PMM3{\interfootnotelinepenalty=10000 \footnote{\url{https://huggingface.co/whaleloops/KEPTlongformer-PMM3}}}. 
Following \cite{Yang2022KnowledgeIP}, we preprocess MIMIC discharge summary by removing de-identification tokens, replacing non-alphanumeric characters (except punctuation) with whitespace, and truncating at 8,192 tokens. 
If the length exceeded this limit, irrelevant sections were removed prior to truncation to retain the most relevant sections.
The MIMIC-III full dataset contains 8,922 ICD codes, making it infeasible to include all descriptions in one prompt. 
We re-rank the top 300 candidates predicted by MSMN \cite{yuan-etal-2022-code} and process 50 candidates at a time, as described in \cite{Yang2022KnowledgeIP, pmlr-v225-kailas23a}. 

To determine the consistency loss weight ($\alpha$), we perform an ablation study on the MIMIC-III-50 dataset by varying $\alpha \in \{0.02, 0.05, 0.1, 0.2\}$.  $\alpha = 0.05$ yields the best performance across all evaluation metrics (see Table~\ref{tab:ablation}). This value is applied to all other experiments, as tuning on the larger MIMIC-III-full dataset is computationally expensive. For the top-50 and rare-code datasets, we randomly select 4 synonyms to construct the augmented prompt $P_a$, following findings from \cite{yuan-etal-2022-code, gomes-etal-2024-accurate} that using 4 or 8 synonyms improves ICD coding model performance.
For MIMIC-III-full dataset, we use the same code descriptions for both $P$ and $P_a$, as we observed that incorporating synonyms increases training time and slows convergence on this larger dataset.

All experiments are conducted on a single NVIDIA H100 80GB GPU, with training time and hyperparameters detailed in Appendix~\ref{appendix:hyperparameter}. We use the Llama-3.1-70B-Instruct model to perform zero-shot acronym expansion for all ICD coding experiments, except for the ablation study reported in Table~\ref{tab:expansion_peformance}.




\section{Results}
\subsection{Acronym expansion performance}
Our experimental results indicate that the size of the large language model (LLM) used for zero-shot prompting has minimal impact on acronym detection. 
Detection precision ranges from 93.7\% with the smallest model to 96.6\% with the largest, while recall increases from 84.5\% to 90.5\%. 
In contrast, model size has a substantial effect on acronym expansion accuracy, showing a notable performance improvement of +42 percentage points, from 18.8\% with the 1B-parameter model to 60.8\% with the 70B-parameter model (Table~\ref{tab:expansion_peformance}). 
Evaluating expansions under the lenient accuracy criterion yields additional gains of up to 4\%.

Table~\ref{tab:abreviation_examples} illustrates example outputs from acronym expansion using the Llama-3.1-70B-Instruct model, along with their corresponding inverse edit distance scores used for lenient accuracy evaluation. 
As shown in the second section of the table, several expansions remain marked as incorrect even under the lenient 70\% similarity threshold, demonstrating our effort to avoid misclassifying incorrect expansions as correct (for example, the first example in the third section). As such, lenient accuracy should be viewed as a conservative, lower-bound estimate of the model’s true acronym expansion performance.

\begin{table*}[h]
\centering
{\fontsize{9}{10}\selectfont  
    \setlength{\tabcolsep}{4pt}  
    \renewcommand{\arraystretch}{1.2}  
\begin{tabular}{l|ccccc|cccc}
\hline
     \multirow{3}{*}{\textbf{Methods}} & \multicolumn{5}{c|}{\textbf{MIMIC-III-50}} & \multicolumn{4}{c}{\textbf{MIMIC-III-full}}\\
\cline{2-10}     
    & \multicolumn{2}{c}{\textbf{AUC}} & \multicolumn{2}{c}{\textbf{F1}} & \textbf{Pre} & \multicolumn{2}{c}{\textbf{F1}} & \multicolumn{2}{c}{\textbf{Pre}}\\
\cline{2-10}
    & \textbf{Macro} & \textbf{Micro} & \textbf{Macro} & \textbf{Micro} & \textbf{P@5} & \textbf{Macro} & \textbf{Micro} & \textbf{P@8} & \textbf{P@15}\\
\hline
    MSMN\cite{yuan-etal-2022-code}                  & 92.8 & 94.7 & 68.3 & 72.5 & 68.0 & 10.3 & 58.4 & 75.2 & 59.9\\
    PLM-ICD\cite{huang-etal-2022-plm}               & - & - & - & - & - & 10.4 & 59.8 & 77.1 & 61.3\\
    DiscNet+RE\cite{zhang-etal-2022-automatic}      & - & - & - & - & - & \textbf{14.0} & 58.8 & 76.5 & 61.1\\
    KEPT (KL)\cite{Yang2022KnowledgeIP} \textdagger               & 92.6 & 94.8 & 68.9 & 72.9 & 67.3 & 11.8 & 59.9 & 77.1 & 61.5\\
    CoRelation \cite{luo-etal-2024-corelation}      & 93.3 & 95.1 & 69.3 & 73.1 & 68.3 & 10.2 & 59.1 & 76.2 & 60.7\\
    MSAM\cite{gomes-etal-2024-accurate}             & 93.7 & \underline{95.4} & \underline{70.4} & \underline{74.0} &\underline{ 68.9} & - & - & - & -\\
\hline
\multicolumn{10}{l}{\textit{Extra human annotations}}\\
\hline
NoteContrast\cite{pmlr-v225-kailas23a} \textdagger         & \underline{93.8} & \underline{95.4} & 69.2 & 73.6 & 68.6 & 11.9 & \underline{60.7} & 77.8 & 62.2\\
MRR \cite{wang-etal-2024-multi} & 92.7 & 94.7 & 68.7 & 73.2 & 68.5 & 11.4 & 60.3 & 77.5 & 62.3 \\
AKIL \cite{wang-etal-2024-auxiliary} & 92.8 & 95.0 & 69.2 & 73.4 & 68.3 & 11.2 & 60.5 & \underline{78.4} & \underline{63.7}\\
\hline
\multicolumn{10}{l}{\textit{Ours}}\\
\hline
    KEPT (KL) \textdagger \textasteriskcentered          & 93.1 & 94.9 & 68.5 & 72.7 & 67.6 & 11.3 & 60.3 & 77.5 & 61.6\\
    KEPT (PMM3) \textdagger \textasteriskcentered    & 93.6 & 95.2 & 69.6 & 73.5 & 68.3 & 12.9 & 61.5 & 78.4 & 62.7 \\
    \textbf{ACE-ICD (KL)} \textdagger & \textbf{93.9} & \textbf{95.6} & \textbf{70.9} & \textbf{74.5} & \textbf{69.2} & 11.7 & \textbf{61.8} & \textbf{78.6} & 63.0\\
   \textbf{ACE-ICD (PMM3) }  \textdagger & \textbf{94.4} & \textbf{95.9} & \textbf{71.6} & \textbf{75.0} & \textbf{70.0} & \underline{13.2} & \textbf{62.7} & \textbf{79.4} & \textbf{63.9}\\
\hline
\multicolumn{10}{l}{\textit{GPT4-based}}\\
\hline
LLM-codex \cite{yang2023surpassing} & 92.9 & 94.8 & 67.4 & 71.5 & - & - & - & - & -\\
Multi-Agents \cite{li2024exploring} & - & - & \textbf{74.8} & 58.9 & - & - & - & - & -\\
\end{tabular}
}
\caption{\label{tab:50_full}
Results on MIMIC-III-50 and MIMIC-III-full datasets, using KEPTLongformer (KL) and KEPT-PMM3 (PMM3). Our reproduced KEPT results (marked with *) closely align with those reported by \cite{Yang2022KnowledgeIP}. Methods marked with \textdagger{} indicate approaches that re-rank the top 300 predictions from MSMN \cite{yuan-etal-2022-code} under the MIMIC-III-full setting.}  
\end{table*}

\subsection{ICD coding performance}

\begin{table}[h]
\centering
{\fontsize{9}{10}\selectfont  
    \setlength{\tabcolsep}{1pt}  
    \renewcommand{\arraystretch}{1.2}  
\begin{tabular}{l|c|cccc}
\hline
    \multirow{2}{*}{\textbf{Methods}} & \textbf{Trained} & \multicolumn{2}{c}{\textbf{AUC}} & \multicolumn{2}{c}{\textbf{F1}} \\
\cline{3-6}
     & \textbf{from} & \textbf{Macro} & \textbf{Micro} & \textbf{Macro} & \textbf{Micro}\\
\hline
    MSMN                 & & 75.4 & 77.4 & 15.3 & 16.6  \\
    KEPT (KL)              & & 79.4 & 80.7 & 24.6 & 23.3  \\
    NoteContrast         & pre-trained & \underline{85.7} & \underline{86.7} & \underline{39.0} & \underline{41.8}  \\
    ACE-ICD (KL)                           & & \textbf{86.8} & \textbf{89.1} & 37.9 & 37.7  \\
    ACE-ICD (PMM3)                           & & \textbf{92.2} & \textbf{90.9} & \textbf{49.1} & \textbf{51.1} \\
    
\hline
    MSMN                & & 59.0 & 58.9 & 3.5 & 5.5 \\
    KEPT  (KL)              & MIMIC & 82.3 & 83.7 & 29.0 & 31.4 \\
    NoteContrast         & III-50 & \underline{88.9} & \underline{89.9} & \underline{40.3} & \underline{42.6} \\
    ACE-ICD (KL)  & checkpoint & \textbf{90.0} & \textbf{90.9} & \textbf{45.3} & \textbf{48.1}  \\
    ACE-ICD (PMM3)   & & \textbf{91.1} & \textbf{91.9} & \textbf{54.0} & \textbf{55.8}\\
\hline
\multicolumn{6}{l}{\textit{GPT4-based}}\\
\hline
LLM-codex           & & 82.5 & 83.2 & 27.9 & 30.2 \\
Multi-Agents         & & - & - & \textbf{71.5} & 37.6 \\
\hline
\end{tabular}
}
\caption{\label{tab:rare}
Results on the MIMIC-III-rare50 dataset.}
\end{table}

Results show that \textbf{ACE-ICD} outperforms previous state-of-the-art methods across all three MIMIC-III datasets (Table \ref{tab:50_full} and \ref{tab:rare}), regardless of whether KEPTLongformer or KEPT-PMM3 is used for initialization.  
From now on, we refer to ACE-ICD (PMM3) as ACE-ICD.
To assess statistical significance, we conduct 1,000 rounds of permutation testing comparing the predictions of ACE-ICD and the KEPT baseline. The resulting p-values are all below 0.05 across evaluation metrics and datasets, indicating that our proposed approach yields statistically significant improvements in ICD coding performance.

On the MIMIC-III-50 task (Table \ref{tab:50_full}), ACE-ICD achieves a macro AUC of 94.4 (+0.6), micro AUC of 95.9 (+0.5), macro F1 of 71.6 (+1.2), and micro F1 of 75.0 (+1.0) and precision@5 of 70.0 (+1.1), with values in parentheses indicate improvements over the previous best results. For the MIMIC-III-full task (Table \ref{tab:50_full}), ACE-ICD outperforms prior methods on most metrics except macro F1, achieving a macro F1 of 13.2 (-0.8), micro F1 of 62.7 (+2.0), precision@8 of 79.4 (+1.0), and precision@15 of 63.9 (+0.2). 
We manage to improve the macro F1 score to 14.3 (+0.3), by applying code-specific threshold optimization (see Appendix \ref{appendix:threshold}).
Under the MIMIC-III-rare50 setting (Table \ref{tab:rare}), ACE-ICD achieves a macro AUC of 91.1 (+2.2), micro AUC of 91.9 (+2.0), macro F1 of 54.0 (+13.7), and micro F1 of 55.8 (+13.2) when fine-tuned from the MIMIC-III-50 checkpoint. 
Notably, ACE-ICD fine-tuned from KEPT-PMM3 outperforms prior methods initialized from the MIMIC-III-50 checkpoint.


\begin{figure*}[t]
  \includegraphics[width=\textwidth]{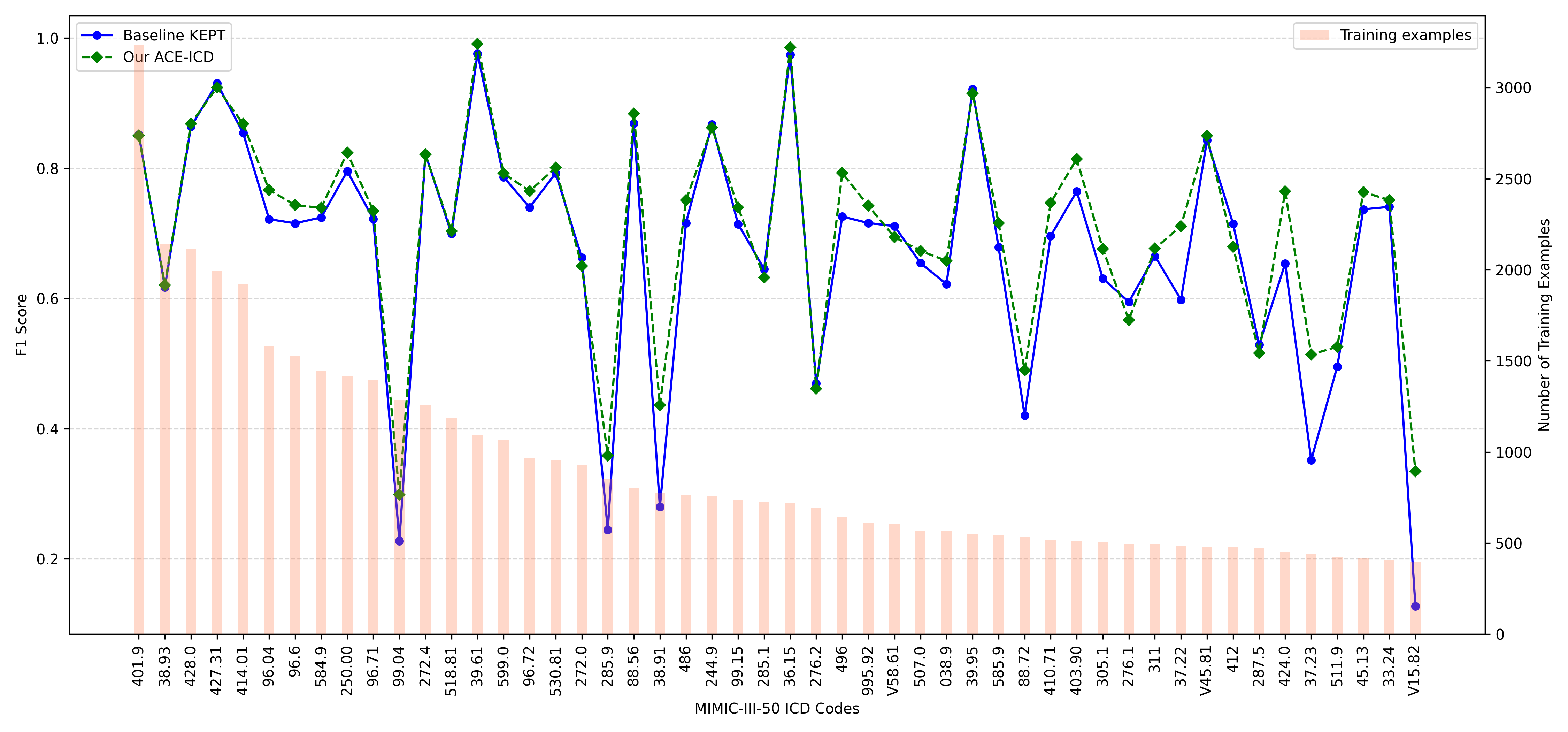}
  \caption{F1 improvement per code in MIMIC-III-50 dataset (sorted by number of training examples in descending order).}
  \label{fig:performance}
\end{figure*}

\section{Discussion}
\begin{table}
\centering
{\fontsize{9}{10}\selectfont  
    \setlength{\tabcolsep}{3pt}  
    \renewcommand{\arraystretch}{1.2}  
\begin{tabular}{l|ccc}
\hline
    \textbf{Model} & \textbf{Macro F1} & \textbf{Micro F1} & \textbf{P@5}\\
\hline
    KEPT (baseline)                          & 69.6 & 73.5 & 68.3\\
    + data augmentation only         & 70.0 & 73.8 & 68.6\\
    + consistency training only           & 71.2 & 74.7 & 69.6\\

    ACE-ICD  (+ both)                   & \textbf{71.6} & \textbf{75.0} &  \textbf{70.0}\\
\hline
    $\alpha$ = 0.2  & 71.0 & 74.6& 69.4\\
    $\alpha$ = 0.1  & \textbf{71.6} & \textbf{75.0}& 69.7\\
    $\alpha$ = 0.05  & \textbf{71.6} & \textbf{75.0} & \textbf{70.0}\\
    $\alpha$ = 0.02   & 71.4 & 74.7 & 69.4\\

\hline
\end{tabular}
}
\caption{\label{tab:ablation}
Ablation study on MIMIC-III-50.}
\end{table}
\quad \textbf{Effectiveness of our proposed training framework.} \quad We evaluate the impact of acronym augmentation and consistency training in ACE-ICD through an ablation study by adding each component separately and assessing performance on the MIMIC-III-50 dataset (Table \ref{tab:ablation}).
For data augmentation only, we include the augmented data into the original dataset, doubling the training set size while halving the number of training epochs.
For consistency training only, we apply R-Drop \cite{wu2021r} directly to the original data.
Both approaches improve performance over the reproduced KEPT baseline, with consistency training yields better gains. 
We attribute this to the fact that zero-shot acronym expansion may introduce translation errors, adding noise to the training data.
In contrast, R-Drop acts as dropout augmentation and has been shown to enhance ICD coding performance by preventing overfitting \cite{yuan-etal-2022-code, luo-etal-2024-corelation}.
However, the performance further improves when both strategies are applied together.

\begin{table*}[h]
\centering
{\fontsize{9}{10}\selectfont  
    \setlength{\tabcolsep}{4pt}  
    \renewcommand{\arraystretch}{1.2}  
\begin{tabular}{l|cccc||ccccc}
\hline
     \multirow{3}{*}{\textbf{Expansion models}} & \multicolumn{4}{c||}{\textbf{Acronym Expansion Performance}} & \multicolumn{5}{c}{\textbf{ICD coding Performance}}\\
\cline{2-10}     
    & \multicolumn{2}{c}{\textbf{Detection}} & \multicolumn{2}{c||}{\textbf{Accuracy}} & \multicolumn{2}{c}{\textbf{AUC}} & \multicolumn{2}{c}{\textbf{F1}} & \multicolumn{1}{c}{\textbf{Pre}}\\
\cline{2-10}
    & \textbf{Precision} & \textbf{Recall} & \textbf{Strict} & \textbf{Lenient}  & \textbf{Macro} & \textbf{Micro} & \textbf{Macro} & \textbf{Micro} & \textbf{P@5}\\
\hline
    Llama-3.2-1B-Instruct & 93.7 & 84.5 & 18.8 & 19.7 & 94.0 & 95.6 & 70.5 & 73.9 & 69.2\\
    Llama-3.2-3B-Instruct & 95.3 & 84.3 & 31.0 & 33.1 & 94.2 & 95.7 & 71.4 & 74.7 & 69.6\\
    Llama-3.1-8B-Instruct & \textbf{96.6} & 86.9 & 46.3 & 49.5 & 94.1 & 95.7 & 71.5 & 74.8 & 69.6\\
    Llama-3.1-70B-Instruct & \textbf{96.6} & \textbf{90.5} & \textbf{60.8} & \textbf{64.7} & \textbf{94.4} & \textbf{95.9} & \textbf{71.6} & \textbf{75.0} & \textbf{70.0} \\
\hline
\multicolumn{5}{c||}{\textit{Without acronym expansion (consistency training only)}} &  94.2 & 95.7 & 71.2 & 74.7 & 69.6 \\ 
\hline
\end{tabular}
}
\caption{\label{tab:expansion_peformance}
Performance of instruction-tuned Llama model variants on acronym expansion and ICD coding. Expansion accuracy is evaluated on the reverse-substituted dataset from \citet{rajkomar2022deciphering}, and ICD coding results are reported on MIMIC-III-50.}
\end{table*}

\textbf{Impact of acronym expansion quality in improving ICD coding performance.} \quad To evaluate how the quality of the acronym expansion process affects ICD coding performance, we conduct an ablation study using Llama 3 models of varying sizes and compare them to a baseline without acronym expansion, as shown in Table \ref{tab:expansion_peformance}. 
We observe that poor expansion quality can hurt performance: the 1B model, with less than 20\% accuracy, degrades coding results. 
Moderate-size models (3B and 8B), achieving 30–50\% accuracy, yield only minor gains.
In contrast, the 70B model, with over 60\% exact match accuracy, provides the most significant performance gains. 
By observing a few incorrectly expanded acronyms illustrated in Table \ref{tab:abreviation_examples}, we hypothesize that expansion errors may have minimal impact if the expanded acronyms are unrelated to any ICD code.

\textbf{Effect of acronym expansion as data augmentation on different ICD codes.} \quad
Figure \ref{fig:performance} illustrates the F1-score improvement for each ICD code, comparing our ACE-ICD model to the baseline KEPT model in the MIMIC-III-50 settings.
The results show that ACE-ICD outperforms the baseline on 39 out of 50 codes, with only marginal decreases observed for the remaining ones.
Our method effectively improves the performance of lower-performing codes, including 99.04 (22.8 $\rightarrow$ 29.9), 285.9 (24.5 $\rightarrow$ 35.9), 38.91 (28.0 $\rightarrow$ 43.6), 37.23 (35.2 $\rightarrow$ 51.4), and V15.82 (12.7 $\rightarrow$ 33.4). 
A Pearson correlation of –0.25 (p = 0.08) between absolute F1 improvement and the number of training examples per code suggests that rarer codes tend to benefit more from our method, although the evidence is not statistically significant.

Moreover, although our approach is designed to improve model robustness to the use of acronyms in clinical texts, the impact of acronym expansion on ICD coding performance varies depending on the presence of code-relevant evidence  in abbreviated or expanded form across different codes. 
If several full-form expressions are already present in the text, acronym expansion may not be necessary for accurate code prediction (e.g., “tka” and “total knee replacement” as evidence for code 81.54 in the example from Table~\ref{tab:comparison}).

A case study is shown in Table \ref{tab:comparison}. Our ACE-ICD demonstrate a better understanding of medical acronyms, correctly predicting codes such as 599.0 and 715.36. 
In this discharge summary (HADM\_ID=108519), the only evidence for inferring code 599.0 is the acronym "uti", as other mentions of "urine" in the text are not relevant. 
Similarly, code 715.36 can be inferred from "oa" or its synonym "osteoarthritis".
Despite the KEPT model being pre-trained to incorporate knowledge from UMLS terms, it still fails to correctly predict these codes, highlighting the effectiveness of our approach in handling medical acronyms.

\quad \textbf{Comparison with recent methods.} \quad Our method outperforms previous state-of-the-art models across all three MIMIC-III datasets, including those using additional human annotations or auxiliary clinical knowledge \cite{pmlr-v225-kailas23a, wang-etal-2024-multi, wang-etal-2024-auxiliary}. 
Notably, our approach significantly improves over the KEPT baseline using a smaller model, KEPTLongformer (149M parameters), and even outperforms MSAM\cite{gomes-etal-2024-accurate}, built on the larger GatorTron model (345M parameters), under the common code setting. 

Even though, GPT-4-based approaches show promising results in ICD coding \cite{yang2023surpassing, li2024exploring}, they still underperform compared to fine-tuned encoder-based models, particularly in terms of micro-F1.
The multi-agent framework proposed by \citet{li2024exploring} achieves a higher macro-F1, which suggests better handling of rare codes due to GPT-4’s extensive medical knowledge, but it exhibits lower micro-F1, reflecting challenges in accurately predicting frequent codes.(See tables \ref{tab:50_full} and \ref{tab:rare}). 
Additionally, these methods require repeatedly sending clinical notes to third-party services to access proprietary LLMs, raising concerns around privacy, cost, and scalability. 
In contrast, our method delivers strong performance on both frequent and rare codes through a lightweight, targeted data augmentation strategy that uses open-source LLMs locally in a one-time preprocessing step, preserving privacy and efficiency.

\section{Related works}
\subsection{Automatic ICD coding}
Automatic ICD coding is a multi-label classification task that assigns diagnosis and procedure codes to clinical notes \cite{perotte2014diagnosis, nguyen2023mimic}. Early approaches use CNNs \cite{mullenbach-etal-2018-explainable, 10.1145/3357384.3357897}, LSTMs \cite{ijcai2020p461, nguyen-etal-2023-two}, and Transformers \cite{huang-etal-2022-plm} to encode clinical notes, while incorporating label attention mechanisms to capture relationships between the notes and ICD codes.
Recent studies further improve code representation by integrating multiple code synonyms \cite{yuan-etal-2022-code, gomes-etal-2024-accurate} or code relation graph learning \cite{luo-etal-2024-corelation}. 
Contrastive learning has also been applied to improve model capabilities, either between medical entities in UMLS \cite{Yang2022KnowledgeIP} or between clinical notes and ICD codes \cite{pmlr-v225-kailas23a}.
Several studies enhance ICD coding performance by leveraging the discourse structure of clinical notes, such as using section type embeddings \cite{zhang-etal-2022-automatic} or contrastive pre-training between sections \cite{lu2023towards}.

While most methods rely solely on the provided clinical notes and code descriptions, some studies focuses on enhancing ICD coding performance using extra human annotations or data augmentation.
\citet{pmlr-v225-kailas23a} pre-train a model on temporal sequences of diagnostic codes using proprietary data from a large patient cohort, where clinical notes are paired with ICD-10 codes, to provide a strong initialization for finetuning coding model.
\citet{wang-etal-2024-multi, wang-etal-2024-auxiliary} incorporate auxiliary information such as diagnosis-related group (DRG) codes, current procedural terminology (CPT) codes, and prescribed medications to improve performance.
\citet{lu2023towards} introduce masked section training with small ratio as a data augmentation strategy, following contrastive pre-training between note's sections to boost model performance.
\citet{falis-etal-2022-horses} propose ontology-guided synonym augmentation and sibling-code replacement to generate silver training examples. However, their method requires a pretrained named entity recognition and linking system to identify code-relevant text spans.

LLMs have demonstrated remarkable capabilities across various general-domain tasks and have recently been explored for ICD coding \cite{boyle2023automated, soroush2024large}. 
\citet{yang2023surpassing} introduced LLM-Codex, which generates ICD codes and evidence with GPT-4, followed by LSTM-based verification.
\citet{li2024exploring} use GPT-4 to convert discharge summaries into Subjective, Objective, Assessment, and Plan (SOAP) format, allowing multiple agents to perform ICD coding via predefined workflows.
While promising, these approaches still lag behind fully fine-tuned non-LLM models on MIMIC-III common and rare datasets.

To the best of our knowledge, no prior work has explored augmenting data with acronym expansions and incorporating them via consistency training to enhance ICD coding performance.
Moreover unlike previous approaches, our approach does not rely on additional annotations or specialized pretraining. Instead, given the evidence of the zero-shot capabilities of general-purpose LLMs to expand medical acronyms, our approach provides a simple yet effective data augmentation strategy to improve ICD coding performance.
\subsection{Clinical acronyms disambiguation}
Accurately disambiguating clinical acronyms and abbreviations enhances automated clinical note processing which include medical information retrieval and analysis. Several studies focus on training deep learning models with large amounts of annotated data, including word-embeddings \cite{jaber2021disambiguating,wu2015clinical}, convolutional neural networks (CNNs) \cite{skreta2021automatically}, fine-tuning transformer-based models such as BioBERT \cite{li2024exploring} and BlueBERT \cite{hosseini2024leveraging}, and fine-tuned encoder-decoder architectures like T5 \cite{rajkomar2022deciphering}.
To address the scarcity of annotated data and the data-hungry nature of deep learning models, prior work has explored various data augmentation strategies, including reverse substitution \cite{rajkomar2022deciphering, liu2024exploring, skreta2021automatically}, UMLS-based similar concept retrieval \cite{skreta2021automatically}, integration of clinical note metadata \cite{kugic2024disambiguation}, and the use of generative clinical models \cite{hosseini2024leveraging}.
The potential of LLMs in medical context understanding, coupled with their reduced reliance on large annotated datasets, has driven research toward zero-shot and few-shot acronym disambiguation in clinical text requiring less training cost and effort.
\citet{kugic2024disambiguation}, \citet{liu2024exploring} and \citet{hosseini2024leveraging} evaluate various LLMs on the CASI dataset \cite{moon2012clinical}, showing that LLM-based prompting achieves performance comparable to supervised models, even in zero-shot settings.
\section{Conclusion}
In this paper, we introduce ACE-ICD, a system that advances ICD coding performance by utilizing acronym expansion as an innovative data augmentation technique. Furthermore, we incorporate consistency training, a regularization strategy that enforces alignment between original and augmented documents to enhance model predictions. Our approach also outperforms studies which rely on external annotations or proprietary resources. Our extensive experiments reveal that the combination of LLM-based acronym expansion and consistency training elevates ICD coding accuracy, outperforming existing methods and establishing new state-of-the-art benchmarks across various settings.

\section*{Limitations}
Our data augmentation approach relies on zero-shot prompting to disambiguate medical acronyms in clinical notes, making its effectiveness dependent on the performance of the selected LLMs. 
We chose Llama 3.1 70B as it was the best-performing open-source model available at the time of our experiments and aligned with our computational resources.
More advanced LLMs or prompting techniques could potentially reduce translation errors and generate higher-quality augmented data.

Our work uses KEPT as the base method, but we argue that acronym expansion as a data augmentation technique, combined with consistency training, can benefit other existing ICD coding systems. Additional experiments are needed to thoroughly assess the effectiveness of our proposed strategy across various models.

\section{Ethics Statement}
This work uses the publicly available MIMIC-III clinical dataset, which contains de-identified patient information in compliance with HIPAA standards. 
Access to the dataset requires completion of a data use agreement and training in responsible research conduct. 
Acronym expansion was performed using open-source LLMs on a secure local cluster, and no patient data were transmitted to any third-party services.
Our method is intended to support clinical NLP research and is not designed for direct clinical deployment without expert oversight. 
We do not anticipate any ethical concerns associated with this study.
\bibliography{custom}
\bibliographystyle{acl_natbib}

\appendix
\section{Appendix}

\subsection{Threshold Optimization}
\label{appendix:threshold}
All results presented in the main paper are obtained using a single threshold for all ICD codes, optimized for the micro F1. 
An alternative approach is to determine a specific threshold for each ICD code. 
This method effectively lowers the threshold for ICD codes with less training data compared to a single-threshold approach, leading to an improvement in the macro F1 and surpassing \cite{zhang-etal-2022-automatic} on the MIMIC-III-full dataset. 
However, this strategy tends to overfit the development set, increasing the performance gap between the development and test sets.

\begin{table}
\centering
{\fontsize{9}{10}\selectfont  
    \setlength{\tabcolsep}{2pt}  
    \renewcommand{\arraystretch}{1.2}  
\begin{tabular}{l|cccc}
\hline
    \multirow{2}{*}{\textbf{Methods}} & \multicolumn{2}{c}{\textbf{Dev F1}} & \multicolumn{2}{c}{\textbf{Test F1}} \\
    
     & \textbf{Macro} & \textbf{Micro} & \textbf{Macro} & \textbf{Micro} \\
\hline
    ACE-ICD (PMM3) & & & & \\
    w/ single threshold                     & 10.5 & 62.9 & 13.2 & \textbf{62.7} \\
    w/ code-specific thresholds              & 14.9 & 66.0 & \textbf{14.3} & \textbf{61.5} \\  
\hline
    DiscNet+RE  & - & - & 14.0 & 58.8 \\

\end{tabular}
}
\caption{
Results of different threshold optimization approaches on MIMIC-III-full dataset.}
\end{table}\label{tab:threshold}

\subsection{More implementation details}\label{appendix:hyperparameter}
\quad \textbf{Pretrained models.}\quad We initialize our model with two pretrained variants provided by \cite{Yang2022KnowledgeIP}.
KEPTLongformer is based on Clinical Longformer \cite{li2023comparative}, while KEPT-PMM3 builds upon RoBERTa-base-PM-M3-Voc-distill \cite{lewis-etal-2020-pretrained}, a distilled variant of RoBERTa-large pre-trained on PubMed, PMC, and MIMIC-III corpus. 
These models adapt the Longformer sparse attention mechanism \cite{beltagy2020longformer} to handle longer sequences and incorporate medical knowledge through contrastive learning. 

\quad \textbf{Training Details.} \quad Table~\ref{tab:hyperparam} summarizes the training hyperparameters for the three MIMIC-III datasets. Training ACE-ICD (PMM3) for 8 epochs on a single NVIDIA H100 GPU takes approximately 25 minutes for MIMIC-III-rare, 7 hours for MIMIC-III-50, and 5 days for MIMIC-III-full.

\quad \textbf{Inference.} \quad
Evaluating 1,573 examples from the development set of MIMIC-III-50 takes approximately 1 minute and 36 seconds on a single H100 GPU, achieving a throughput of around 16 examples per second. For MIMIC-III-full, the model requires six runs to re-rank 300 candidates, resulting in a throughput of 2.67 examples per second. 
\begin{table*}[t]
\centering
{
\fontsize{9}{10}\selectfont  
    \setlength{\tabcolsep}{6pt}  
    \renewcommand{\arraystretch}{1.2}  
\begin{tabular}{lccc}
     \textbf{Configuration} & \textbf{MIMIC-III-50} & \textbf{MIMIC-III-rare50} & \textbf{MIMIC-III-full} \\
\hline
    global attention on & code descriptions + masks & code descriptions + masks & code descriptions + masks \\
    global attention stride & 1 & 1 & 3\\
    synonyms in prompt & yes & yes & no \\
    max length & 8192 & 8192 & 8192 \\
    num epochs & 8 & 8 & 4\\
    batch size & 1 & 1 & 1\\
    gradient accumulation steps & 1 & 1 & 6\\
    learning rate & 1.5e-5 & 1.5e-5 & 1.5e-5\\
    learning rate scheduler & cosine & cosine & cosine \\
    max grad norm & 1 & 1& 1\\
    warm up ratio & 0 & 0 & 0.1 \\
    AdamW epsilon & 1e-6 & 1e-6 & 1e-7 \\
    AdamW betas & (0.9, 0.999) & (0.9, 0.999) & (0.9, 0.999) \\
    weight decay & 0.01 & 0.01 & 1e-4 \\

\end{tabular}
}
\caption{
Training hyperparameters used in our experiments for the three ICD coding tasks on the MIMIC-III dataset.}
\label{tab:hyperparam}
\end{table*}

\end{document}